\documentclass{article}
\usepackage{authblk}
\usepackage{amsmath}
\usepackage{amsfonts}
\usepackage{amssymb}
\usepackage{url}
\usepackage[table,xcdraw]{xcolor}

\usepackage{graphicx}

\begin{document}
%
%
%
\title{Ultra-Quantisation: Efficient Embedding \\Search via 1.58-bit Encodings}
\author{Richard Connor}
\author{Alan Dearle}
\author{Ben Claydon}
\affil{School of Computer Science, University of St Andrews,\\
St Andrews, Scotland, UK\\
\vspace{5pt}
\{rchc,al,bc89\}@st-andrews.ac.uk}
\date{}

%
%

%
\maketitle              
\begin{abstract}

Many modern search domains comprise high-dimensional vectors of floating point numbers derived from neural networks, in the form of \textit{embeddings}. 
Typical embeddings range in size from hundreds to thousands of dimensions, making the size of the embeddings, and the speed of comparison, a significant issue.

Quantisation is a class of mechanism which replaces the floating point values with a smaller representation, for example a short integer. This gives an approximation of the embedding space  in return for a smaller data representation and a faster comparison function.

Here we take this idea almost to its extreme: we show how vectors of arbitrary-precision floating point values can be replaced by vectors whose elements are drawn from the set $\{-1,0,1\}$. This yields very significant savings in space and metric evaluation cost, while maintaining a strong correlation for similarity measurements.

This is achieved by way of a class of convex polytopes which exist in the high-dimensional space.
In this article we give an outline description of these objects, and show how they can be used for the basis of such radical quantisation while maintaining a surprising degree of accuracy.

\end{abstract}
{\textbf{keywords}: high-dimensional search, quantisation, ternary quantisation, 1.58 bit quantisation, convex polytope.}
\section{Introduction and context}

\begin{table}
\caption{Table of notations used}
\label{table:notations}
\begin{tabular}{|p{.3\textwidth}|p{.7\textwidth}|} \hline 

{\itshape Notation} & {\itshape Context of use}\\ \hline 

 $(U,\delta)$&a universal search space with distance function $\delta$\\ \hline  
 $(S,\delta)$&a large finite search space\\ \hline  
 $\{s \in S : \delta(q,s) \le t_k \}$&the knn search result for some query $q_k \in U$. Note that the knn threshold $t_k$ depends on $q$.\\ \hline 
 $\ell_2^d$&The Euclidean distance over $d$ dimensions; $d$ is elided where clear from context\\ \hline  
 $(U',\delta')$&A proxy space which allows faster processing than $(U,\delta)$ and gives correlated judgements\\ \hline  
 $(\mathbb{R}^d,\ell_2)$&a $d$-dimensional Euclidean space\\ \hline  
 $\mathbb{T}$&the ternary set of integers $\{-1,0,1\}$\\ \hline  
 $(\mathbb{B}^{2d},\Delta)$&a proxy space for $(\mathbb{R}^d,\ell_2)$, replacing each element of $\mathbb{R}$ with 2 bits and $\ell_2$ with a very fast function $\Delta$\\ \hline  
$\{x,d\}$ EVP&A family of convex polytopes defined on a $d$-dimensional hypersphere; $x$ is the number of non-zero elements\\ \hline
 $u_i, u_j, \dots$ & elements of the space $(\mathbb{R},\ell_2)$\\\hline
 $v_i, v_j, \dots$ &elements of the proxy space $(\mathbb{T}^d,\ell_2)$\\\hline
 
\end{tabular}
\end{table}

\subsection{Background: Cosine and hyperspherical spaces}

Our context is similarity search over high-dimensional vector spaces deriving from neural network embeddings. We concentrate on $k$-nearest neighbour search (kNN). For some universal search space $(U,\delta)$, where $\delta$ is a dissimilarity metric, we wish to search a large finite set $S \subset U$ with respect to a query $q \in U$. The magnitude of $S$ makes this challenging, otherwise there is no problem to solve. To overcome this, we  typically resort to approximate search techniques, and perform pre-processing of $S$ before $q$ becomes available.

The metric used over these spaces is typically Euclidean or Cosine distance; the choice is relatively unimportant, as in high dimensions the rank order implied by either is similar. We restrict our domain of interest to Cosine (angular) spaces. In the context of kNN applications, the most important property of the dissimilarity metric $\delta$ is the rank ordering it imposes upon $S$ with respect to $q$. In this context, we note a well-known mapping between Cosine space and $\ell_2$-normalised Euclidean space. It is clear that, as Cosine distance measures the angles among vectors, this is unchanged if all vectors are adjusted so that their length (norm) is set to 1. However this allows the space to be viewed as a Euclidean space without affecting the rank order.
We elaborate this point as our analysis  relies upon the  Euclidean geometry of these \textit{hyperspherical} spaces.

One further consequence of applying  $\ell_2$-normalisation is the correlation between Euclidean distance and the negative scalar product: the Euclidean distance $\ell_2(X,Y) =\sqrt{\sum_i (x_i - y_i)^2}$ 
can be rewritten as $\sqrt{\sum_i (x_i^2 + y_i^2 - 2 x_i y_i)}$. Since $\sum_i x_i = \sum_i y_i = 1$ in a $l_2$ normed space this 
is perfectly correlated with -$\sum x_i y_i$. 
This is generally well known by practitioners as an optimisation; here we introduce it explicitly as, in our geometric derivations, we make use of this correlation.

\subsection{Proxy spaces}

One well-known family of pre-processing techniques maps from a search space $(U,\delta)$ to a \textit{proxy} space $(U',\delta')$. The intent is that the proxy space  preserves ranking judgements, and processing over the proxy space $(U',\delta')$ is more efficient than for $(U,\delta)$. Such mappings may be exact or approximate.

Here we introduce an approximate mapping over hyperspherical search spaces.
The universal space is thus $(\mathbb{R}^d,\ell_2)$ with the additional constraint that for each $u \in \mathbb{R}^d, ||u|| = 1$. We show how such a space can mapped into a proxy space $(\mathbb{T}^{d},\ell_2)$ where $\mathbb{T}$ is the (ternary) set $\{-1,0,1\}$, therefore giving an information-theoretic space requirement of only 1.58 bits per element.
This mapping can be achieved with a remarkably small loss of correlation over distances in $(\mathbb{R}^d,\ell_2)$.
Separately, we show how this ternary space can be mapped without loss into a space $(\mathbb{B}^{2d},\Delta)$ where $\mathbb{B}^{2d}$ is the set of binary sequences of length $2d$, for a distance metric $\Delta$ which is very cheap to evaluate. 



The mapping to ternary values is based on the geometry of convex polytopes.
In this article, we will give an outline of the geometry involved, and then concentrate on the pragmatic use of the results in the domain of search.
While the geometry of high-dimensional Euclidean spaces is perhaps less than straightforward, the pragmatic outcomes are simple and easily applicable by other search practitioners.

\subsection{Contribution of this work}

This paper shows a  high-dimensional  geometric basis for a mapping from vectors of floating point numbers to vertices of a high-dimensional polytope, where distances are well preserved. The practical outcome is simple, and shows:

\begin{enumerate}
    \item a simple mapping from high-dimensional floating point vectors to sequences of bits, where each floating point value is replaced by only two bits; and
    \item a fast, SIMD-sympathetic, dissimilarity function over the bit sequences, where
    \item the correlation between this function and distances in the original Euclidean space are very close.
\end{enumerate}

Advantages are in both space and execution speed; while it is difficult to disentangle these in practice, as the smaller memory footprint also gives greater speed, we have measured the combination to give a speedup of over 100 times in typical contexts. At the same time, the Spearman correlation over distances may be as good as 0.96. As the mapping and distance function are entirely generic, they may be used in conjunction with any modern vector search mechanism.

\section{Equi-Voronoi Polytopes}

\subsection{Voronoi centres as proxies}

In outline, our mapping is defined as follows. Within a given Euclidean space, we identify a specific convex polytope. The vertices of this polytope are used to define a Voronoi partition of the hypersphere, with the property that all the cells of the partition have equal volume. 

\begin{figure}
    \centering
    \includegraphics[width=0.5\linewidth]{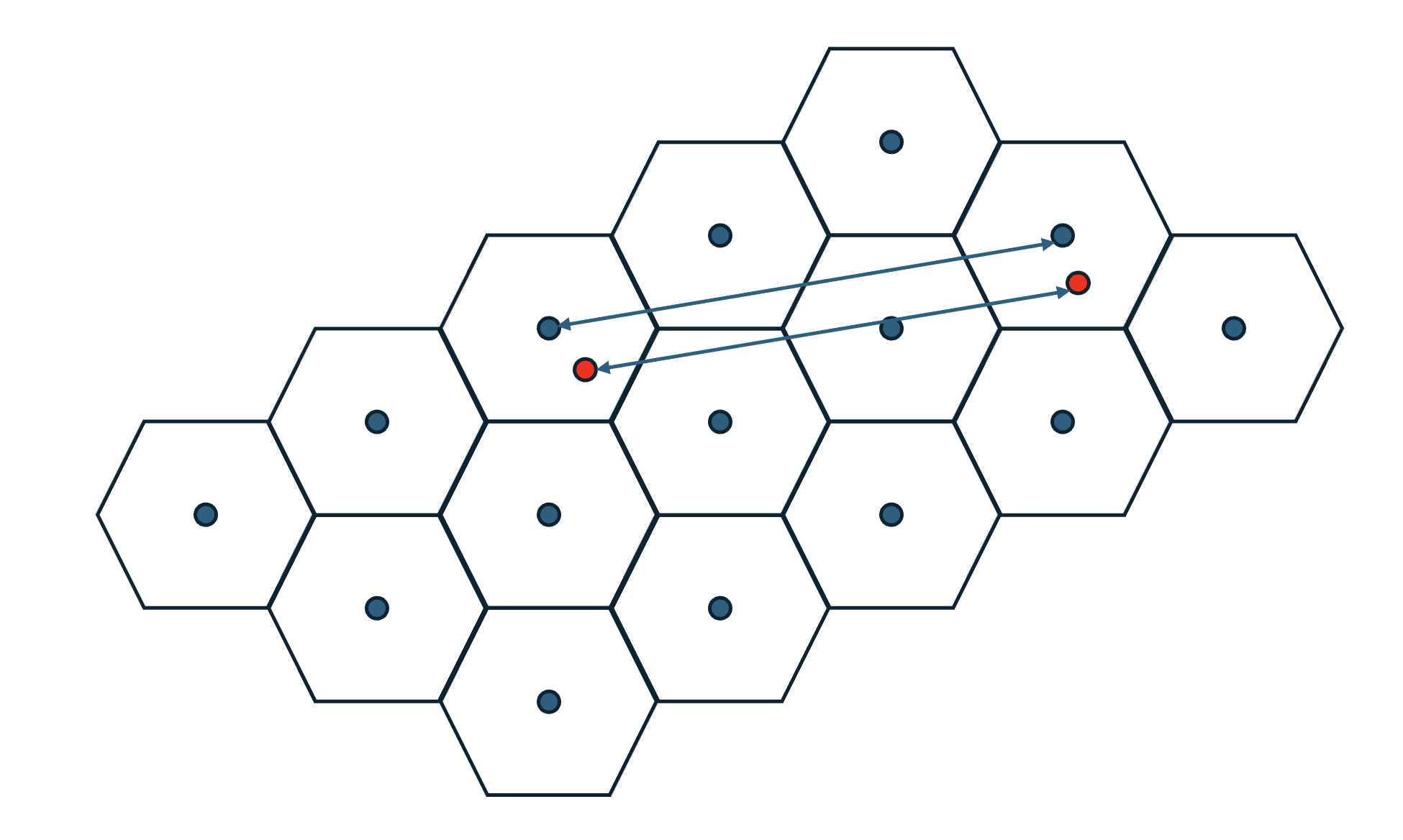}
    \caption{A Voronoi partition in 2D based on a regular set of points. Each cell of the partition comprises that part of the space which is closest to a particular point. In this case, the distance between the centres of their Voronoi cells acts as a good proxy for the distance between the coloured points shown.}
    \label{fig:voronoi_centres}
\end{figure}

Due to geometric constraints in the high-dimensional space, the distance between two vertices acts as a good proxy for the distance between any two points within their respective cells.
Figure \ref{fig:voronoi_centres} shows an example in two dimensions to  illustrate the principle.

The polytope has the properties that:
\begin{enumerate}
    \item its vertices give a very good proxy distance for elements of their cells,
    \item determining the nearest vertex to an arbitrary element is very cheap, and
    \item calculating the distance between vertices is also very cheap.
\end{enumerate}

\subsection{Geometric basis}


We introduce a family of high-dimensional convex polytopes which we refer to as \textit{Equi-Voronoi Polytopes} (EVPs). Each vertex of such an object has the following properties: it has coordinate elements drawn from $\{-1,0,1\}$; it resides on the surface of a hypersphere centred around the origin, and it is at the centre of a equi-volume Voronoi cell with respect to that hypersphere.



\subsection{The EVP vertex space}

For any dimension $d$, and any strictly positive $x \le d$, we define the $\{x,d\}$ EVP as follows:

\begin{itemize}
    \item Each vertex is defined by a $d$-dimensional vector from the origin
    \item Each vector element is one of $\{-1,0,1\}$
    \item Each vector contains precisely $x$ non-zero elements
    \item The $\{x,d\}$ EVP is the convex polytope defined by the set of vertices derived from all possible permutations obtained by following the above rules.
\end{itemize}

\begin{figure}
    \centering
    \begin{minipage}[t]{0.45\textwidth}
        \centering
        \raisebox{20ex}{
        \fbox{
        \begin{tabular}{lll}
        (-1,-1,0) & (-1,0,-1) & (0,-1,-1) \\
        (-1,1,0)  & (-1,0,1) & (0,-1,1)    \\
        (1,-1,0)  & (1,0,-1)  & (0,1,-1)   \\
        (1,1,0)   & (1,0,1)  & (0,1,1)    
        \end{tabular}
        }
        }
    \end{minipage}
    \begin{minipage}[t]{0.45\textwidth}
        \centering
        \raisebox{-3ex}{
        \fbox{
        \includegraphics[width=\linewidth]{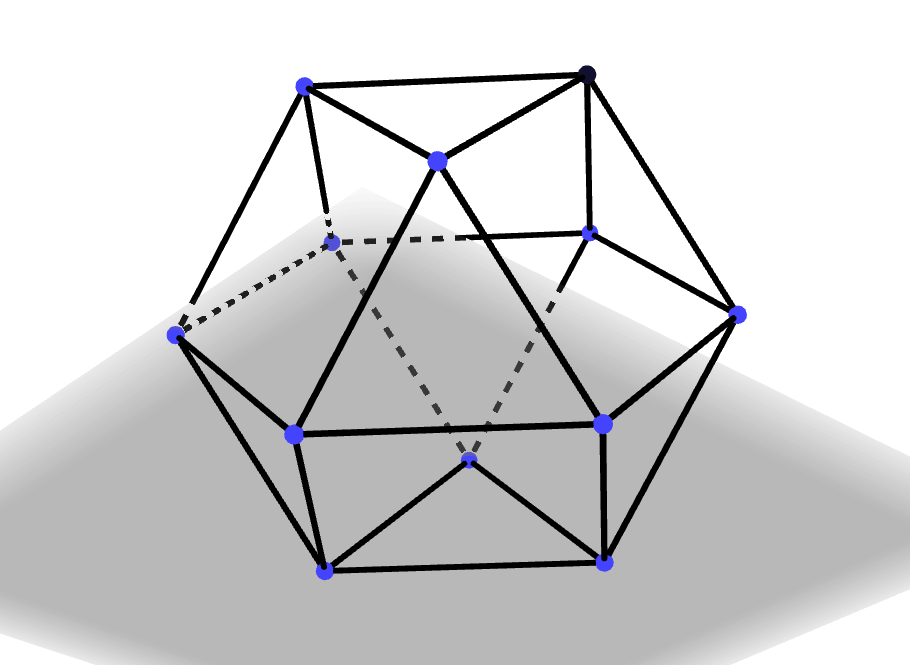}
        }
        }
    \end{minipage}
    \caption{The vertices and 3D form of a cuboctahedron, which is a $\{2,3\}$ EVP. As can be seen, the faces are not uniform. However, considering the sphere containing the vertices, the Voronoi partitions corresponding to each are congruent.}
    \label{fig:cuboct_pic}
\end{figure}

To give a simple example, the $\{2,3\}$ EVP is a structure known as a cuboctahedron. Its vertices and 3D form are shown in Figure \ref{fig:cuboct_pic}.


From the above construction it can be seen that an $\{x,d\}$ EVP has ${d \choose x} \cdot 2^x$ vertices. High-dimensional EVPs have a very large number of vertices; for example, the $\{256,384\}$ EVP, which we will use in Section \ref{sec:experiments},  has around $10^{180}$ vertices. 

\subsection{Mapping to the vertex space}

Our observation is that, if high-dimensional vectors of floating point numbers are assigned to their nearest EVP vertex as a proxy, then the proxy distances among these vertices give a surprisingly accurate approximation of the true values.

Assume for simplicity that the space being mapped, and the EVP vertex space, are set within a hypersphere of the same diameter.
The Euclidean distance between two points is then perfectly correlated with their negative scalar product. The nearest $\{x,d\}$ EVP vertex to an arbitrary  point in the hyperspherical space is therefore that which maximises the scalar product of the vectors representing the two points.


The nearest vertex 
can therefore be constructed by the following steps:

\begin{enumerate}
    \item find the $x$ largest absolute elements in the vector representing the datum
    \item assign 1 to every positive element within this set, and -1 to every negative element
    \item assign 0 to all other elements
\end{enumerate}

A simple example of this mapping is included in Table \ref{tab:float_bit_translations} in Section \ref{sec:dist_measurement}.

\subsection{Equi-Volume Voronoi cells}

The mapping given above to the nearest vertex can also be used to demonstrate that, considering the Voronoi partition defined by the set of all EVP vertices within the hypersphere, each has the same volume. 

For any set of vectors describing uniformly distributed points on the surface of a hypersphere, each vector element is necessarily drawn from an identical and independent distribution \cite{voelker2017}. From this observation it is clear that, for an arbitrarily selected point from a uniform distribution, its nearest vertex from the set of all possible vertices is equiprobable. As the number of points selected tends to infinity, the number associated with each vertex therefore becomes closer to equal, implying that the volume of each cell is the same.

\subsection{Calculating vertex distance}

Every vector in the vertex space has the same magnitude, as the number of non-zero elements is fixed, and therefore the negative scalar product can be used as a proxy for Euclidean distance between values.
To calculate the scalar product of two vectors, each of whose elements is drawn from $\{-1,0,1\}$,
can be a much faster computation than the comparison of vectors whose elements are arbitrary floating point numbers. 

We give more detail in Section \ref{sec:experiments}. For now, we note that:

\begin{itemize}
    \item the limited range of values in the vertex vectors imply that no actual multiplication is necessary to calculate the scalar product: only addition is required, and
    \item a further mapping to a pair of bit vectors, where bits are used to represent the locations of the values 1 and -1, allow a bitwise operation to be used to calculate the scalar product.
\end{itemize}

The experiments in Section \ref{sec:experiments} show this calculation can be one or two orders of magnitude faster on commodity hardware than the scalar product of floating point values. However the smaller memory footprint required also implies that these benefits may be magnified in real search applications when many high-dimensional values need to be stored, copied, and loaded in a relatively restricted memory.

\section{Geometric constraints}

The remaining question is why the EVP vertex distance should give a very good estimate of the true distance. To explain this we  consider  four points: $u_i$ and $u_j$ from the universal space, and $v_i$ and $v_j$ from an EVP vertex space, where the latter are the nearest vertices to the former.

\begin{figure}
    \centering
    \includegraphics[width=0.45\linewidth]{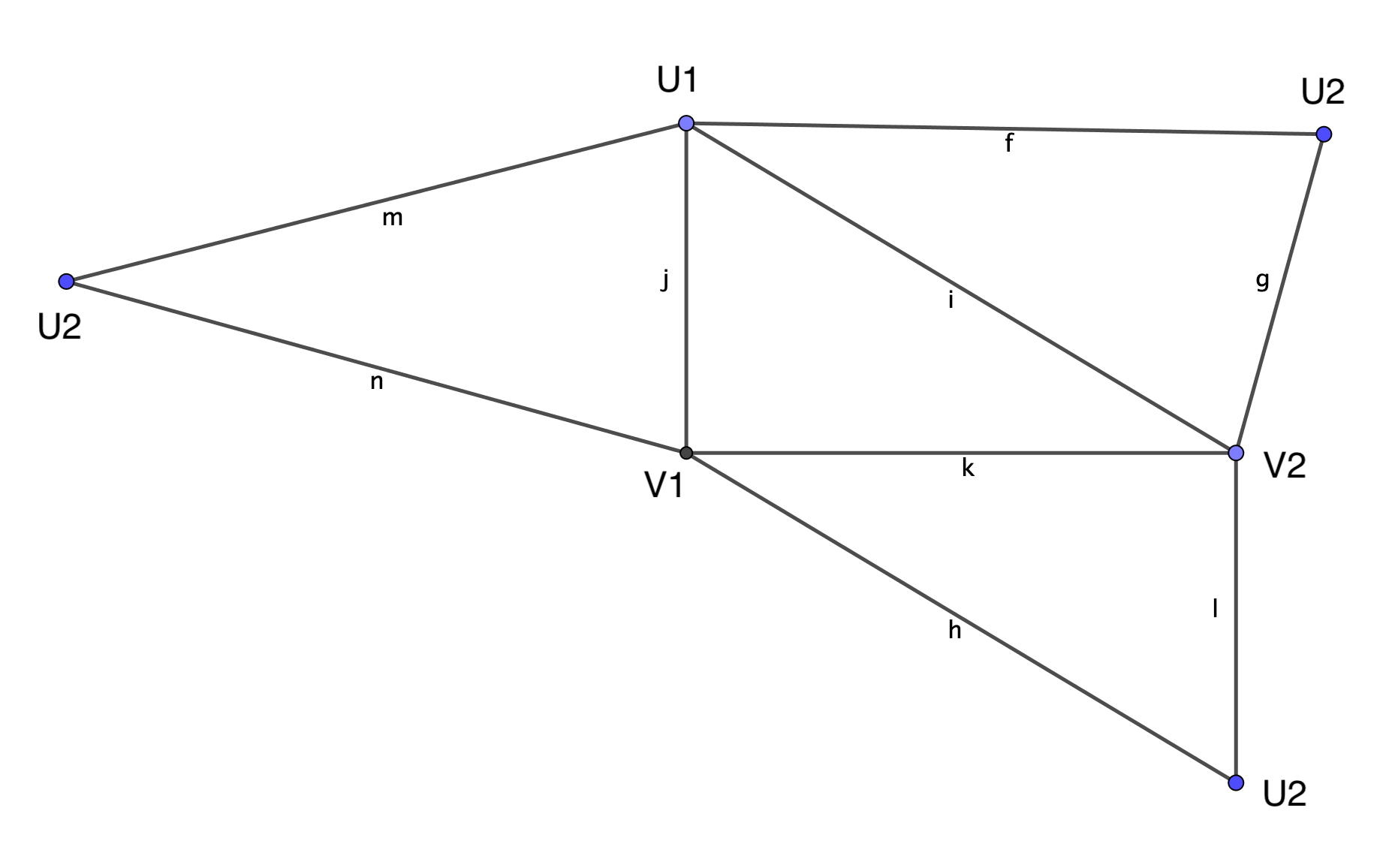}
    \includegraphics[width=0.45\linewidth]{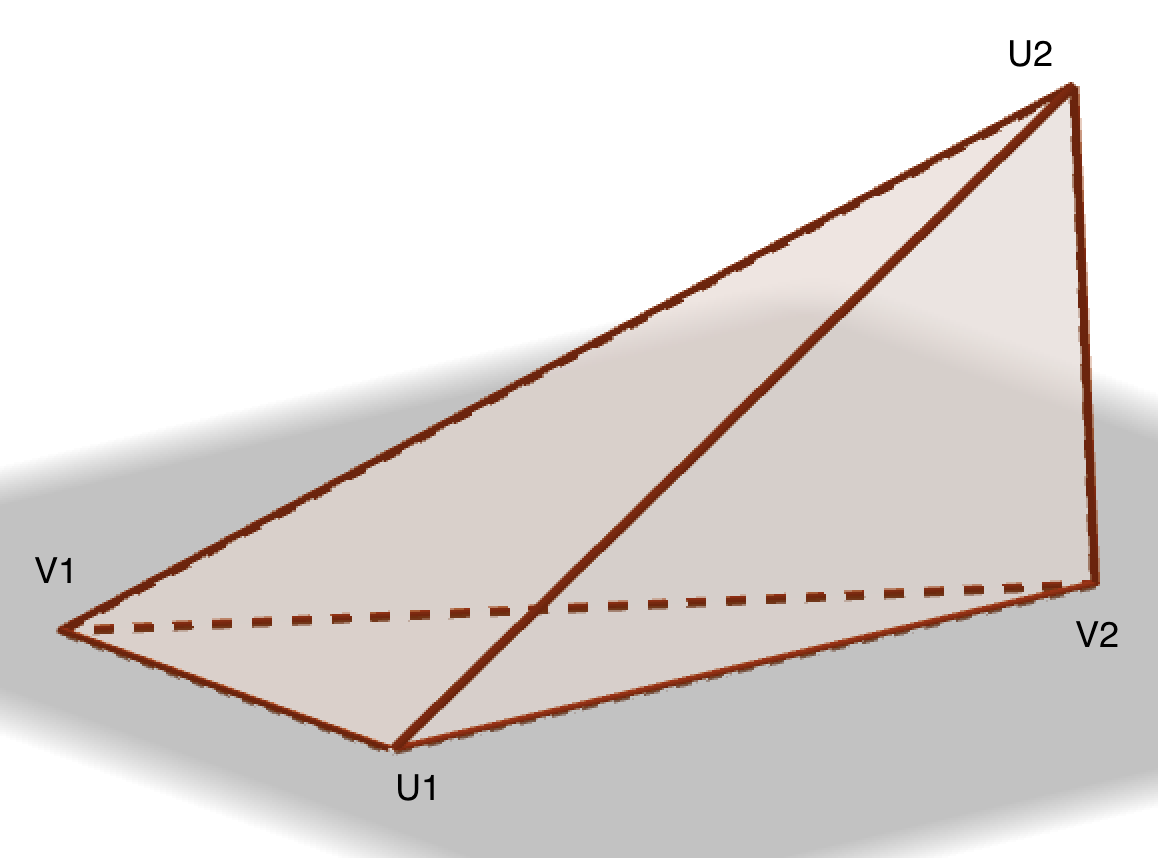}
    \caption{A tetrahedral net and its 3D rendering based only on the distance $\ell_2(V_1,V_2)$. From high-dimensional geometry we can assume: (1) the sides $V_1 U_1$ and $V_2 U_2$ are almost the same length, which is less than $V_1 V_2$; the angles $\angle V_1 V_2 U_2$ and $\angle V_2 V_1 U_1$ are almost right angles; and (3) the angle between the triangles with the common edge $V_1 V_2$ is also close to a right angle. These constraints leave only a little possible variance in the length of the unknown side $U_1 U_2$. }
    \label{fig:tetrahedra}
\end{figure}

The full geometric explanation for the phenomenon is complex, and we refer the interested reader to a fuller exposition \cite{evp_in_prep} we give elsewhere. An outline explanation follows:

\begin{enumerate}
    \item The ``surface'' of a high-dimensional hypersphere is, in its own right, a high-dimensional Euclidean volume%
     \footnote{The terms ``surface'' and  ``volume'' are only correctly used for objects in three dimensions. As here however, the terms are commonly used in higher dimensions to distinguish relative geometric domains.}. 
    For example the surface of a hypersphere set within 250 dimensions is a 249-dimensional space \cite{blumenthal1953}.

    \item The Voronoi cells defined by the EVP vertices are fully enclosed structures in which the EVP vertex is central. The dimensionality of the space implies that almost the entire volume of each cell is at the furthest possible distance from the Voronoi centre \cite{Blum_Hopcroft_Kannan_2020}. For example, a hypersphere has a volume of $V_1r^d$, where $V_1$ is the volume of the unit sphere, $r$ is the radius, and $d$ is the dimensionality of the space. To quantify this, a unit hypersphere in 250 dimensions contains only one-millionth of its volume within a radius of 0.95. It is therefore the case that, for any pair of two points comprising $u_i$ and its corresponding nearest vertex $v_i$, the distance $\ell_2(u_i,v_i)$ is almost the same.

    \item The ``surface'' space possesses the 4-point property: that is, the distances among any four points can be used to construct an isometric tetrahedron in 3D space \cite{blumenthal1933note,connor35hilbert}. We can therefore consider our four points of interest, $v_i,v_j,u_i$ and $u_j$ as the vertices of a simple tetrahedron whose edge lengths preserve their pairwise distances in the high-dimensional space. A tetrahedron constructed from these four points is shown in Figure \ref{fig:tetrahedra}.

    \item We now consider angles within that tetrahedron; in particular, the angle $\angle v_i,v_j,u_j$. It is well known that, in a high-dimensional Euclidean space, a pair of randomly sampled vectors are very likely to be close to orthogonal \cite{Blum_Hopcroft_Kannan_2020}. This result was extended in \cite{sampled_angles} to the angles formed within triangles defined by triples of selected points%
    \footnote{Such triangles are always defined in any proper metric space.}.
    As the distance $\ell_2(v_j,u_j)$ is relatively small, the high dimensions of the space results in a very restricted range for angle $\angle v_i,v_j,u_j$. This is due to the apparent density of space within the hypersphere as viewed from a fixed viewpoint: it is tightly gathered close to the equator of the hypersphere as viewed from a pole%
\footnote{The volume within a $d$-dimensional space, along the line segment $v_1,v_2$, is $\int_{-r}^{r} C \sin^{(d-1)} (\cos^{-1} 1/r) dr$ \cite{sampled_angles}. For large $d$, $\sin^d \theta$ is close to 0 if $\theta \not \approx \pi/2 $.}. 
By symmetry, the same observations are true of the angle $\angle v_j,v_i,u_i$. 
Together with (3), this starts to give some quite tight probabilistic lower and upper bounds for $\ell_2(u_i,u_j)$.

    \item Finally, we consider the set of all tetrahedra containing the two triangles with the common edge $\overline{v_i v_j}$.  These can be described by rotation of these triangles around their common edge between 0 and $2\pi$.   As explained in \cite{connor2024nsimplex}, this angle is also subject to similar angular constraints of the high-dimensional space, and is again very likely to be close to a right angle, again with a small variance.
\end{enumerate}
\noindent
The angular constraints of the last two points now give a very restricted range of likely values for the distance $\ell_2(u_i,u_j)$.  Figure \ref{fig:tetrahedra} shows a tetrahedral net and its 3D rendering constructed in the light of the above observations to illustrate the point.

\section{Related Work}

The general notion of replacing high-precision values with lower precision is well known.
For high-dimensional vectors, high precision floating point values usually carry more information than necessary. If space is not an issue, double precision (64-bit) floating point values are often used, but it is generally the case that coercion into 32-bit or even 16-bit (half-precision) floating point values loses little information. 

One issue with quantised values is that hardware support for operations such as scalar product, or more general matrix multiplication, is slow to evolve. After demand is evident, it is necessary for hardware providers to build support into processors, followed by library (eg BLAS) standardisation changes, followed by language support for compilation into these libraries. For example, while half-precision arithmetic was standardised in 2002, it became available via BLAS on NVIDIA's Volta architecture only in 2017, and standard operations such as matrix multiplication are still not available for this format in many language environments. A major advantage of our ternary representation, and its 2-bit encoding, is that the $b_2sp$ function, given in Section \ref{sec:dist_measurement}, can be made very fast on standard SIMD hardware.

\subsection{8-bit integer quantisation}
\label{eight_bit_quant}
Floating point values may be converted to 8-bit integer values through a simple linear conversion. There is necessarily some loss of information, but in high-dimensional vectors this is minimal. Although a space saving is achieved, once again hardware-supported arithmetic is not possible without access to specialist hardware, such as Google's tensor processing units whose primary purpose is to support matrix multiplication over short integers.

\subsection{1-bit quantisation}
\label{sec:one_bit_quant}

One-bit quantisation is the extreme form, and is achieved simply by replacing positive values by 1 and negative values by 0. This gives the largest possible space saving, with the added benefit that Hamming distance may be used in place of Euclidean. In the domain of similarity search, this is an equivalent representation to that of sketches \cite{mic2019binary} formed from the polar extremes of each dimension.

It is interesting to note that, for a vector in $d$ dimensions, the 1-bit quantised form is equivalent to finding the nearest vertices of the $\{d,d\}$ EVP.

\subsection{b1.58 quantisation}
\label{sec:158_bit_quant}

Recently, the use of ternary quantisation has emerged in large language model research \cite{ma2024era, ma2025bitnet}. The main observation is that weight vectors in neural networks may be quantised into this domain to give an excellent tradeoff between memory footprint and quality for activations of the model. The quantisation function used is based on the mean of the absolute values in the weight matrix $W$:
\[
\gamma = \frac{1}{nm} \sum_{i,j}{|W_{ij}|}
\]
and the quantised form is then given by
\[
\widetilde{W} = \max(-1,\min(1,\text{round}\left(\frac{W}{\gamma + \epsilon}\right)))
\]
The authors state that a theoretical model explaining their results is  unknown. We observe that, in many cases, this quantisation function gives a matrix which is fairly similar to those of the EVP vertices representing its columns, and we suspect our  theory presented here explains their observations.

In Section \ref{sec:experiments}, we compare our EVP vertex representation with the 1-bit and 1.58-bit encodings described above.

\section{Experimental analysis}
\label{sec:experiments}
In this Section, we present experimental results to justify three hypotheses:

\begin{description}
    \item[Hypothesis 1] Distance correlations between general vector spaces and their corresponding EVP vertex spaces are tight
    \item[Hypothesis 2] The EVP representation can be used to give good search results
    \item[Hypothesis 3] The scalar product of EVP vertices can be calculated very quickly on commodity hardware
\end{description}

\subsection{Methodology}

We use the following data sources:

 
\begin{enumerate}
    \item sets of generated (uniformly distributed on the hypersphere) vectors, in 100 and 1,000 dimensions
    \item  GloVe-100  data \cite{glove} (100 dimensions), as taken from the ANN benchmark site \cite{ann_benchmarks} 
    \item PubMed data (384 dimensions) taken from the SISAP Indexing Challenge site \cite{sisap_challenge}
    \item CLIP data \cite{clip_laion} (768 dimensions) taken from the Laion data set \cite{sisap_challenge}. The CLIP data was reduced without any loss to 500 dimensions using PCA, and this was followed by a random rotation to avoid the principal components being tightly aligned with the EVP vertices.
\end{enumerate}

All data sets have around one million elements; other attributes are summarised in Table \ref{tab:data_etc}. Code for the experiments was written in MatLab and Rust, and is available from \cite{}.

\begin{table}[]
    \centering
\caption{Data sets used in analysis}
\label{tab:data_etc}
    \begin{tabular}{|c|l|l|c|l|l|l|}\hline
 data set&dimensions&EVP used&no. of vertices & \multicolumn{3}{|c|}{Spearman $\rho$ correlation}\\\hline
 \multicolumn{4}{|c|}{}& EVP& 1-bit&b1.58\\\hline
 Uniform& 100& $\{67,100\}$&$10^{46}$& 0.80& 0.70&0.75\\\hline
 Uniform& 1000& $\{667,1000\}$& $10^{475}$& 0.79& 0.64&0.71\\\hline
 GloVe& 100& $\{67,100\}$& $10^{46}$& 0.83& 0.67&0.75\\\hline
          PubMed& 384&$\{256,384\}$&   $10^{182}$& 0.94& 0.84&0.91\\\hline
          Laion& 500&$\{333,500\}$& 
     $10^{237}$& 0.96& 0.89&0.94\\ \hline\end{tabular}

\end{table}

\subsubsection{Selecting $x$ for a given $d$}

The first task, given a vector space of $d$ dimensions, is to choose a value for $x$ with which to construct the associated $\{x,d\}$ EVP vertex coordinate for each vector. Here we select the value of $x$ which maximises the number of vertices in the EVP, as given by the term ${d \choose x} \cdot 2^x$.  While $2^x$ increases with $x$, $d \choose x$ reduces after $x \ge d/2$.  The whole term can be differentiated using the $\Gamma$ function to create a continuous form, from which it turns out that the inflection point (which always corresponds to the maximum number of vertices) occurs when $x = 2/3 d$.  We have used this EVP form for all experiments.

\subsubsection{Distance measurements}
\label{sec:dist_measurement}

Table \ref{tab:float_bit_translations} shows an example translation from a 10-dimensional Euclidean space to the vertices of a $\{5,10\}$ EVP.

\begin{table}   
\caption{Two floating-point vectors $u_1,u_2$ and their translation to $\{5,10\}$ EVP vertices $v_1,v_2$. For each vector, the five largest absolute values are set to 1 or -1 according to their sign. $v_1^+$ and $v_1^-$ are binary representations whose bits are set only if the corresponding element in $v_1$ is 1 or -1 respectively; similarly for $v_2$. }
    \label{tab:float_bit_translations}
    \centering
    \begin{tabular}{|l||l|c|c|c|c|c|c|c|c|c|c|}\hline
          $u_1$ &float&\cellcolor{lightgray} 0.32&  \cellcolor{lightgray} 0.4&  \cellcolor{lightgray} -0.38&  -0.19&  0.29&  \cellcolor{lightgray} 0.45&  \cellcolor{lightgray} 0.44&  -0.16&  0.23& -0.02
\\\hline
          $u_2$ &float&-0.16&  \cellcolor{lightgray} -0.4&  \cellcolor{lightgray} 0.38&  \cellcolor{lightgray} 0.45&  0.14&  0.19&  \cellcolor{lightgray} -0.38&  -0.04&  \cellcolor{lightgray}0.4& -0.35
\\\hline\hline
          $v_1$ &ternary&\cellcolor{green!70!yellow!40}1&  \cellcolor{green!70!yellow!40}1&  \cellcolor{red!70!yellow!40}-1&  0&  0&  \cellcolor{green!70!yellow!40}1&  \cellcolor{green!70!yellow!40}1&  0&  0& 0
\\\hline
  $v_2$ &ternary&0& \cellcolor{red!70!yellow!40}-1& \cellcolor{green!70!yellow!40}1& \cellcolor{green!70!yellow!40}1& 0& 0& \cellcolor{red!70!yellow!40}-1& 0& \cellcolor{red!70!yellow!40}-1&0
\\\hline \hline
 $v_1^+$ &binary& \cellcolor{green!70!yellow!40}1& \cellcolor{green!70!yellow!40}1& 0& 0& 0& \cellcolor{green!70!yellow!40}1& \cellcolor{green!70!yellow!40}1& 0& 0&0\\\hline
 $v_1^-$ &binary& 0& 0& \cellcolor{red!70!yellow!40}1& 0& 0& 0& 0& 0& 0&0\\\hline \hline
 $v_2^+$ &binary& 0& 0& \cellcolor{green!70!yellow!40}1& \cellcolor{green!70!yellow!40}1& 0& 0& 0& 0& 0&0\\\hline
 $v_2^-$ &binary& 0& \cellcolor{red!70!yellow!40}1& 0& 0& 0& 0& \cellcolor{red!70!yellow!40}1& 0& \cellcolor{red!70!yellow!40}1&0\\\hline
    \end{tabular}

\end{table}

We show that, due to the restricted domain, the value can be calculated very efficiently on commodity hardware, and in particular without performing any multiplication operations.

To achieve this, consider the binary vectors $v_1^+$ and $v_1^-$. These are constructed by simply mapping the position of the values 1 and -1 respectively in the vector $v_1$.  We first note that the scalar product of $v_1$ and any other vector $w$ can be calculated using masked addition: that is, by adding the values of $w$ where $v_1^+ = 1$, and subtracting the values of $w$ where $v_1^- = 1$.

However we can improve further by comparing only these bitstrings.

We note the bitwise scalar product of two binary vectors $bsp(v,w)$ can be calculated by counting the bits of $v \land w$, a very fast operation where all of the operations are amenable to being performed in parallel on a SIMD architecture.

We can then define the function $b_2sp(v,w)$ to give the scalar product of two ternary vectors, based on their binary forms $v^+,v^-,w^+$ and $w^-$:
 \begin{align}
b_2sp(v,w) \quad=\quad &(bsp(v^+,w^+) + bsp(v^-,w^-)) \notag\\
                     &- \quad (bsp(v^+,w^-) + bsp(v^-,w^+)) \notag
 \end{align}
We have found this function to give the best performance outcome in terms of the combination of small space requirements and maximum use of SIMD hardware.

\subsection{Hypothesis 1: Correlation with true distance}

To measure this correlation, sets of value pairs were sampled and the distances between the original vectors and their corresponding nearest vertices measured.

\begin{figure}
    \centering
    \includegraphics[width=0.32\linewidth]{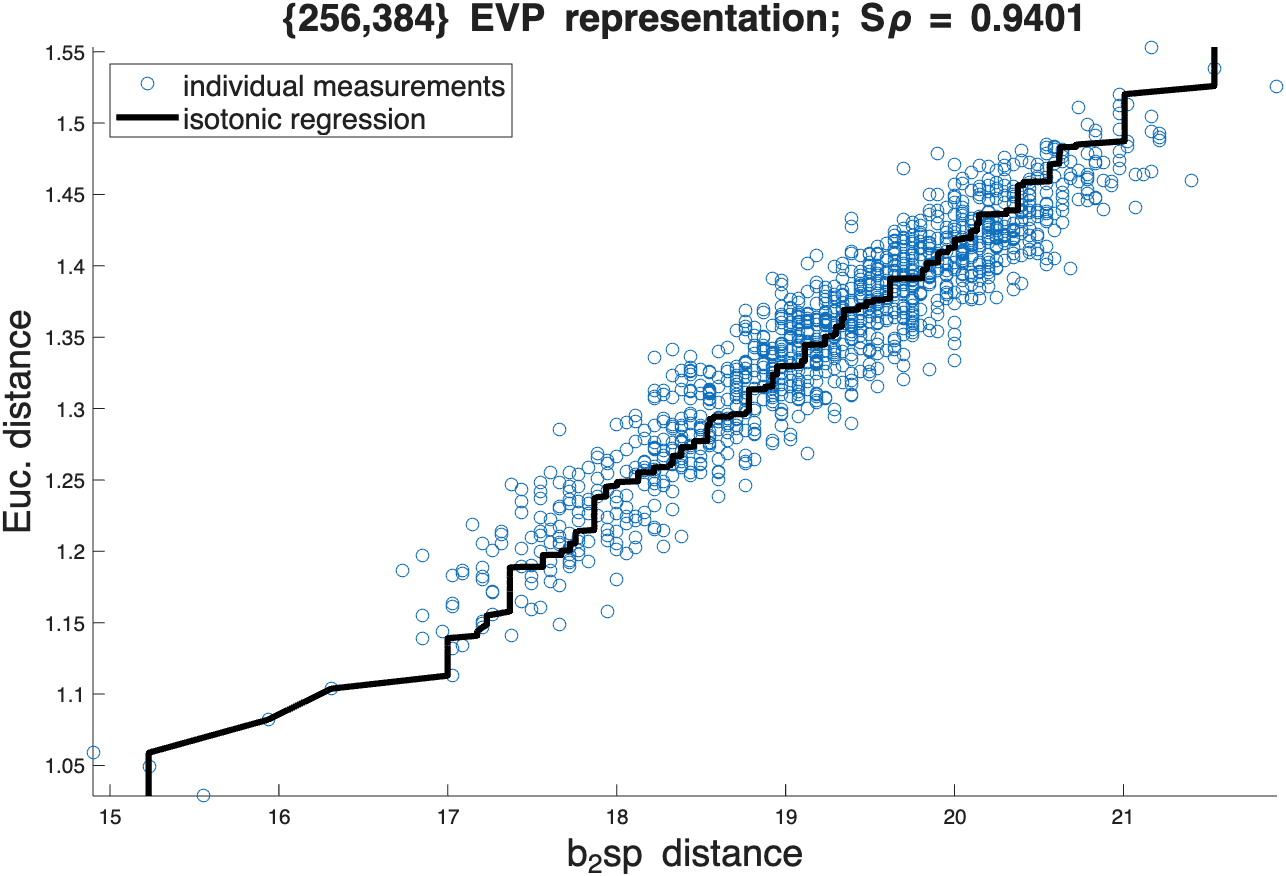}
    \includegraphics[width=0.32\linewidth]{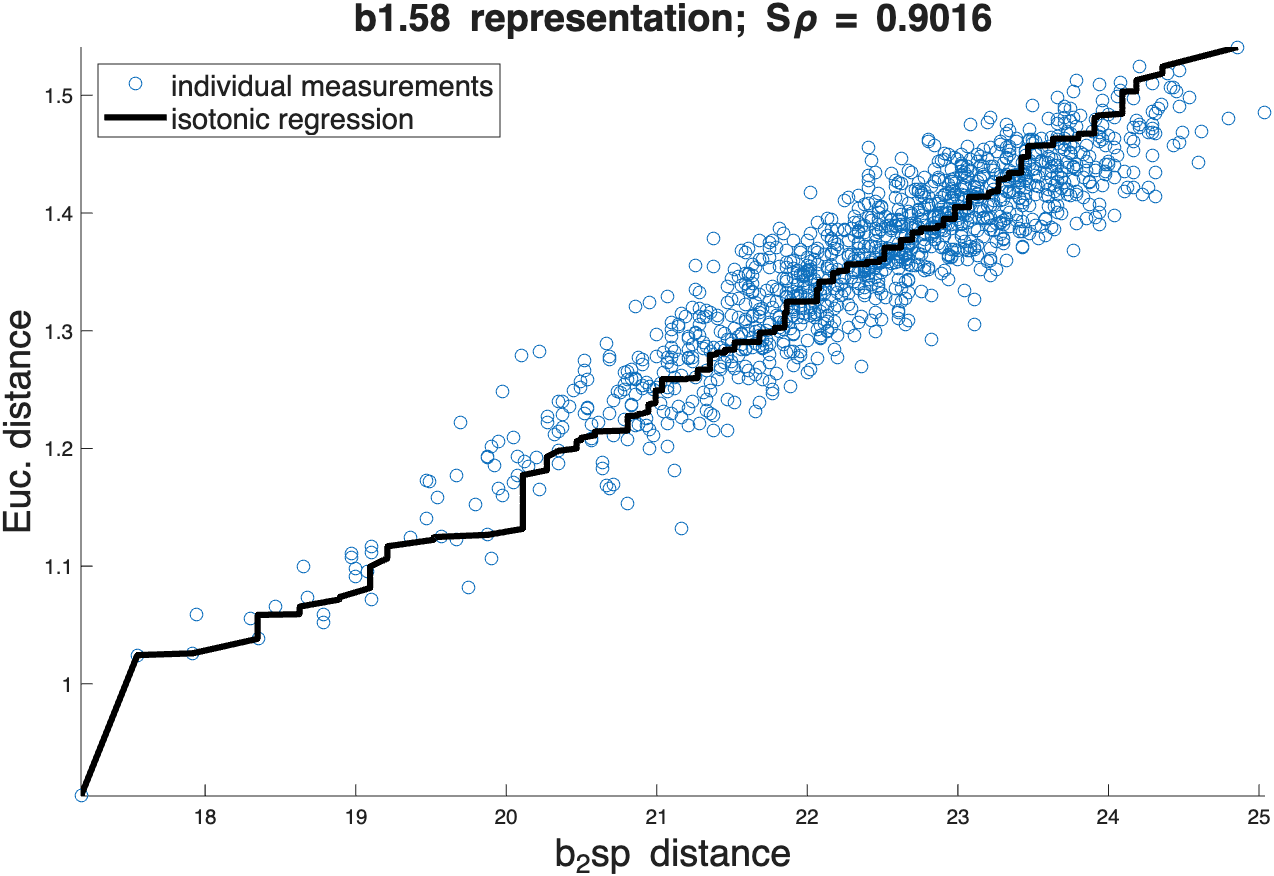}
    \includegraphics[width=0.32\linewidth]{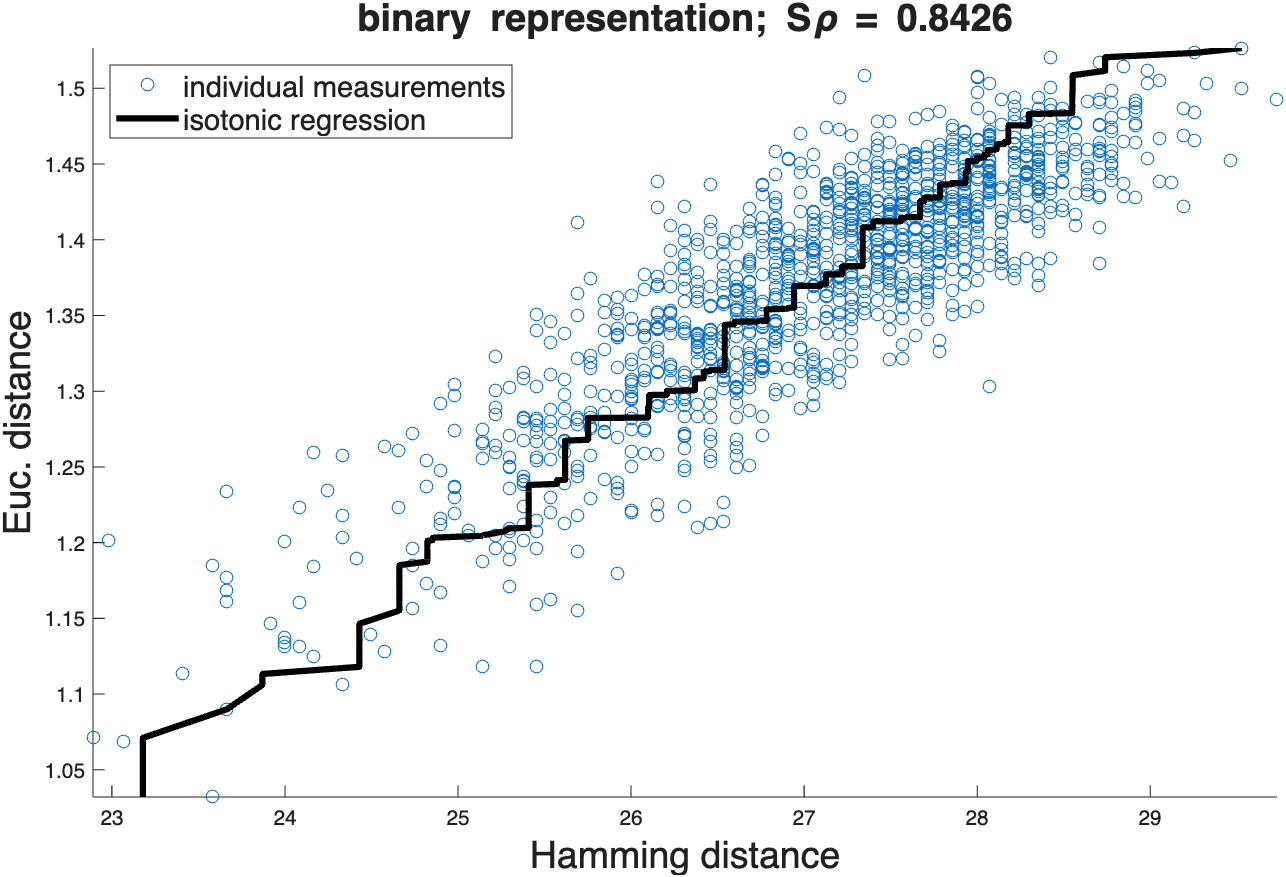}
    \caption{Shepard diagrams showing correlations for randomly selected pairs of points from PubMed. Each plot shows the quantised distance approximation on the X axis against the true distances on the Y axis. The quality of the approximation may be judged by the amount of horizontal displacement from the isotonic regression, which is accurately measured by the Spearman $\rho$ correlation.}
    \label{fig:pubmed_correlations}
\end{figure}

A useful visual impression of distance correlation is given by  Shepard diagrams \cite{dim_red_analysis} which plot the true distance against the approximate distance, along with an isotonic regression of these points and the Spearman-$\rho$ correlation. Figure \ref{fig:pubmed_correlations} shows Shepard diagrams for EVP,  b1.58 and 1-bit quantisations over the PubMed data set. The other data sets give similar visual outcomes.

As can be seen, the EVP quantisation is the best. Table \ref{tab:data_etc} gives the Spearman $\rho$ correlations measured for the three quantisation techniques over all the datasets. It can be seen that, without exception,  the EVP approximation is  best.

\subsection{Hypothesis 2: Search}

Search using approximate techniques often includes a post-processing phase to improve the outcomes. That is, to achieve a kNN search, often   $n > k$ results are found from the approximate search, which are then reordered with respect to the original space, with the best $k$ then selected.

To measure this, we have conducted a range of 30@$n$ searches, where $n$ varies between 30 and 500, over the ``real world'' data sets. For each data set, we measure the fraction  of  30 true near-neighbours which are contained within the closest $n$ neighbours of the proxy space.

\begin{figure}
    \centering
    \includegraphics[width=0.45\linewidth]{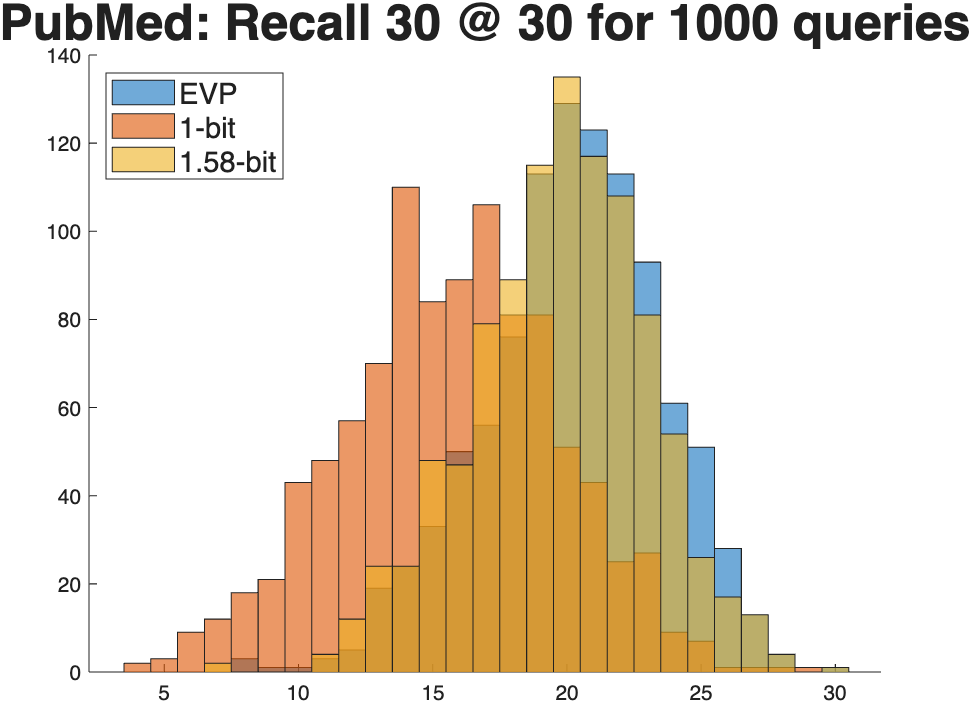}
    \includegraphics[width=0.45\linewidth]{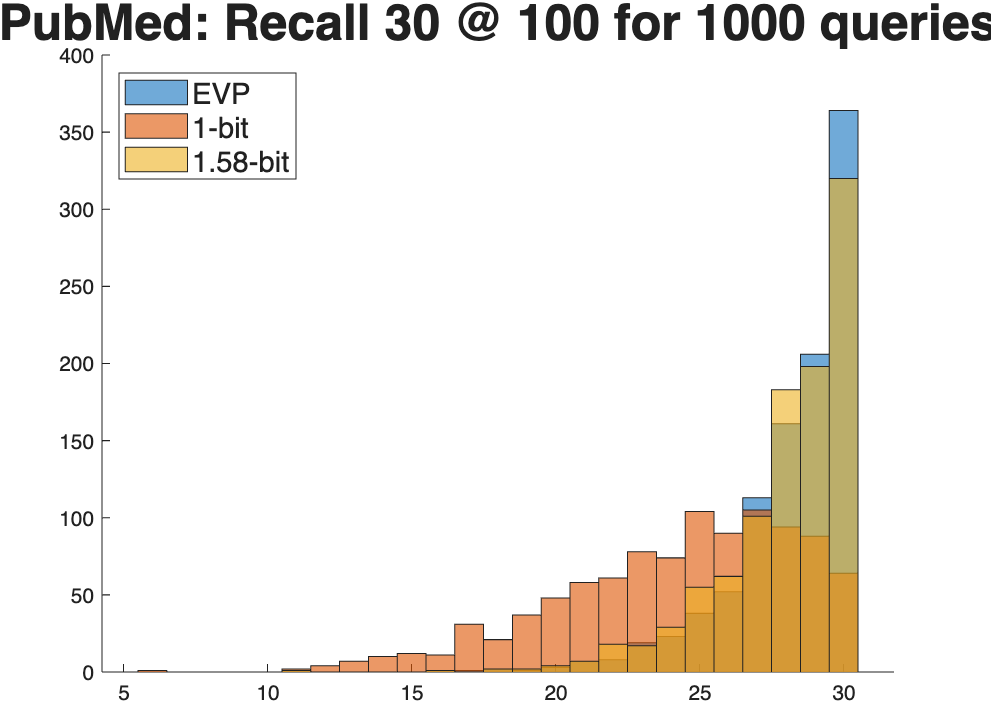}
    \caption{Histograms of 30@30 and 30@100 recall for PubMed data}
    \label{fig:pubmed_search}
\end{figure}

Figure \ref{fig:pubmed_search} shows histograms of the individual recall measured for 1000 queries over the PubMed data set. The left hand figure shows 30@30, and the right hand shows 30@100, for EVP, b1.58 and 1-bit quantisation.

It can be seen in both cases that the EVP quantisation is substantially better than the 1-bit quantisation.  However the difference is dramatically improved between the 30@30 and 30@100 queries. We believe this is due to the geometric constraints of the EVP proxy space.

\begin{figure}
    \centering
    \includegraphics[width=0.31\linewidth]{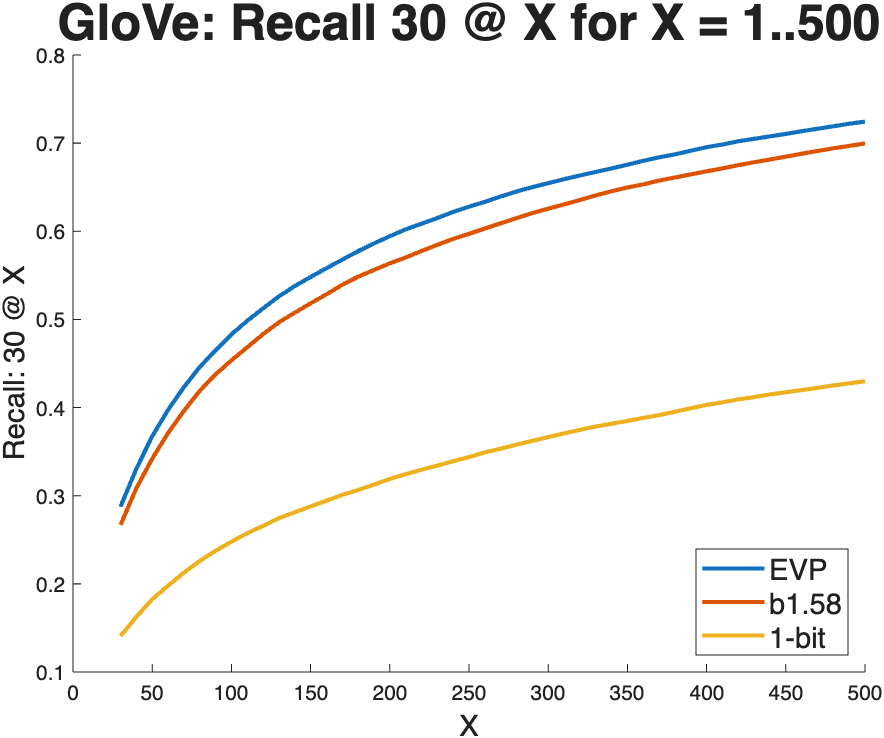}
    \includegraphics[width=0.31\linewidth]{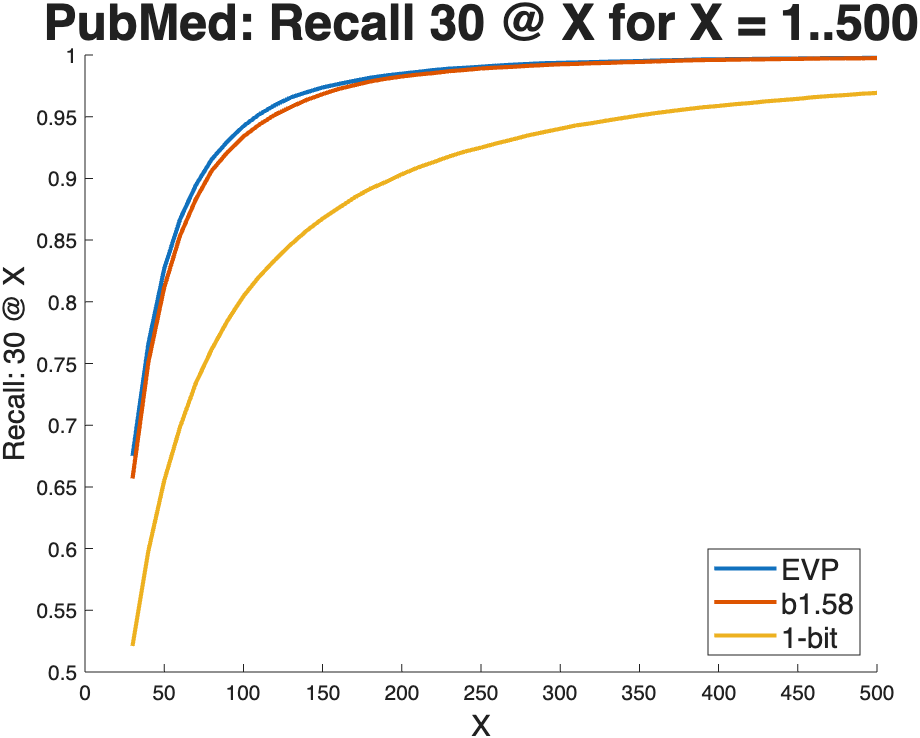}
    \includegraphics[width=0.31\linewidth]{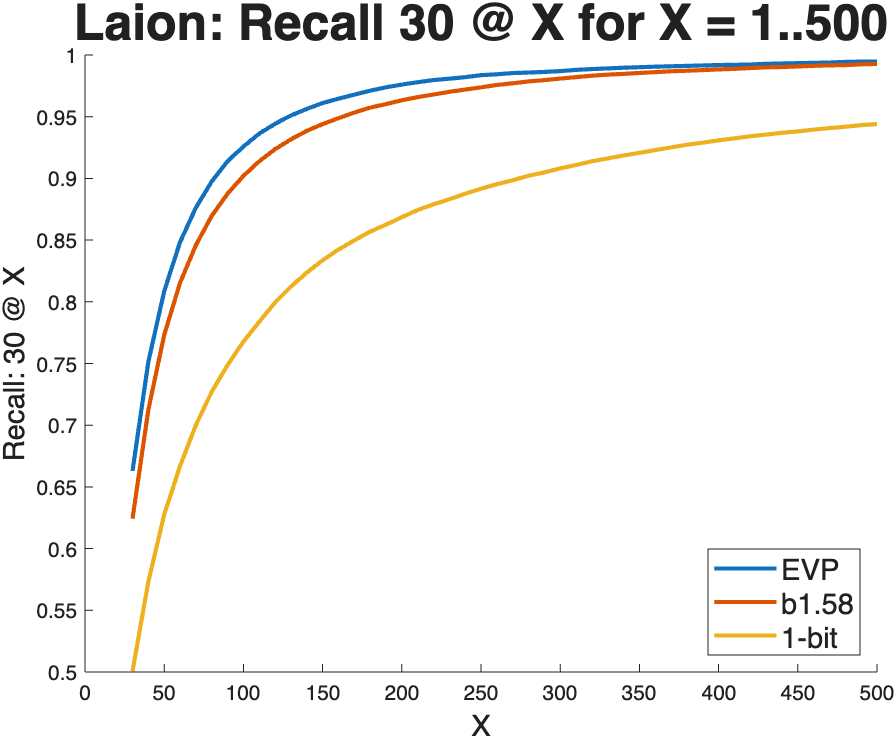}
    \caption{Search results over the real-world data sets}
    \label{fig:thirty_at_n_search}
\end{figure}

Figure \ref{fig:thirty_at_n_search} shows the same experiment repeated for the three real-world data sets with recall values between 30@30 and 30@500.

\subsubsection{Hypothesis 3: Speed of comparison}

Finally, we give some performance measurements comparing the space and speed of the original floating-point calculations and their bitwise equivalents.

We examine the performance in two forms:  using micro-benchmarks that measure the raw performance of similarity judgements, and performing multiple different  calculations over real datasets.

To efficiently encode the $\{x,d\}$ EVP representations as bit vectors, we use the standard Rust data type \texttt{BitVecSimd}%
\footnote{\url{https://docs.rs/bitvec_simd/latest/bitvec_simd/index.html}}.

Each EVP value is represented a pair of \texttt{BitVecSimd} bit vectors whose sizes are rounded up to a multiple of 256 bits. One bit vector represents the $v^+$ bits, the other the $v^-$ bits as described in Section \ref{sec:dist_measurement}.
Thus for an embedding containing $X$ floating point values the number of bits in the bit vector representation is the smallest integer $d$ such that $d * 2 * 256 > 2X$.

For the 384-element PubMed vectors, for example, the bit representations are 1024 bits long. which is 0.08 of the space occupied by 32-bit floating point values.
 
As shown in Section \ref{sec:dist_measurement}, the $b_2sp$ calculation between two $\{x,d\}$ EVP representations may be performed using only logical and arithmetic operations in a SIMD fashion, which is very fast on modern hardware. Table \ref{tab:micro-bechmarks} shows the time taken to execute one million micro-benchmark queries written in Rust. 
The first row gives times when performing a $b_2sp$ operation over the bitmaps. The second row shows the time taken using standard Euclidean distance metric. As can be seen in the figure the distance calculations are more than an order of magnitude faster.

\begin{table}
\caption{Micro-benchmark speeds for binary scalar product and Euclidean distances over 384 dimensions. Times are for 1 million measurements
on an Apple M3 Macbook. The absolute cost per comparison of $b_2sp$ is only around 20 nanoseconds.}
\label{tab:micro-bechmarks}
\centering
\begin{tabular}{|l|c|c|c|c|}
\hline
metric         & fastest   & slowest  & median   & mean\\
\hline
$b_2sp$&  21.61 ms & 21.88 ms & 21.75 ms & 21.76 ms \\ \hline
Euclidean&  258 ms   & 280.3 ms & 263.5 ms & 263.9 ms  \\
\hline
\end{tabular}
\end{table}

Such micro-benchmarks do not tell the whole story.
In particular, they do not measure the benefits incurred when loading and storing the smaller representations through the memory hierarchy.
To investigate the real savings derived from the use of bit encoded $\{x,d\}$ EVP representations we measure end-to-end performance using the Glove, PubMed and Laion data sets, by performing a set of 100 exhaustive queries over the whole data.

Table \ref{tab:compare-metrics} shows  the time taken in milliseconds to run 100 queries,
along with the speed-up resulting from the use of the EVP binary form. As can be seen, the higher dimensional data benefits from more speed-up than the relatively low-dimensional GloVe data. However, even in the case of GloVe a speed up of thirty-three times is achieved.

\begin{table}
\caption{Time in milliseconds to run 100 exhaustive queries over 1M data}
\label{tab:compare-metrics}
\centering
\begin{tabular}{|l|l|l|l|}\hline
               & Glove       & Dino    & Laion \\\hline
Dimensions     & 100         & 384          & 768 \\\hline
Bits           & 200         & 768          & 1536      \\\hline
$\{x,d\}$ EVP& $\{67,100\}$& $\{256,384\}$& $\{512,768\}$\\\hline
Euclidean time & 76.3& 267.3& 564.6\\\hline
$b_2sp$ time& 2.3& 2.6& 35.7\\\hline %
speed up       & 33X         & 103X         & 158X           \\ \hline\end{tabular}
\end{table}

\section{Conclusions and further work}

Here we have presented a mechanism which allows vectors of floating-point numbers to be translated, via ternary values, into pairs of bitstrings where the bitwise scalar products can be used as a proxy for Euclidean distance. This translation is based on a principled geometric analysis of the underlying high-dimensional space, which guarantees a surprisingly close correlation in the resulting distances.

There are a number of other possibilities based on the same geometry which we are currently investigating.

\subsection{Single-operand translation}

In this paper we have concentrated exclusively on similarity measurements between query and data both translated into the proxy space $(\mathbb{B}^{2d},b_2sp)$. We note however that the ``masked addition'' function outlined in Section \ref{sec:dist_measurement} can be applied when only one of the vectors has been translated to ternary form. Early experiments show that comparisons of the ternary form with the original floating-point space, or the same space quantised to 8-bit integers, give an even better accuracy and can be applied to a number of different contexts.

\subsection{Matrix multiplication}

We have so far used the scalar product approximation only as a proxy function for Euclidean similarity. The more general observation that, in a general matrix multiplication, the outcome of each cell in the result matrix is a scalar product over one row and one column of the parameter matrices. It may be that the EVP geometry gives a valuable mechanism for approximate - multiplication free - matrix multiplication.

\subsection{Different EVPs}

All of our experiments here use an EVP construction which maximises the number of Voronoi cells, which generally gives the best search results. We have looked at using much smaller numbers of non-zero elements; the tradeoff is that accuracy decreases, but the speed of masked addition increases. It may that when most of the values in the representation are zero, there are further very fast possibilities for calculating a reasonably accurate approximation.



\bibliographystyle{splncs04}
\bibliography{references,refs} 

\end{document}